\documentclass[10pt,twocolumn,letterpaper]{article}

\usepackage[letterpaper, margin=1.8cm]{geometry}
\usepackage[utf8]{inputenc}
\usepackage[T1]{fontenc}
\usepackage{times}
\usepackage[pagebackref,breaklinks,colorlinks]{hyperref}
\usepackage{url}
\usepackage{booktabs}
\usepackage{float}
\usepackage{graphicx}
\usepackage{tikz}
\usepackage{orcidlink}
\usepackage{authblk}
\usepackage{xspace}
\usepackage{multirow}
\usepackage{caption}
\usepackage{subcaption}
\usepackage{amsmath}

\usepackage{xcolor}
\definecolor{first_rgb}{RGB}{255, 184, 184}
\definecolor{second_rgb}{RGB}{255, 222, 184}
\definecolor{third_rgb}{RGB}{255, 255, 184}
\newcommand*{\first}[1]{{\setlength{\fboxsep}{0pt}\colorbox{first_rgb}{{#1}}}}
\newcommand*{\second}[1]{{\setlength{\fboxsep}{0pt}\colorbox{second_rgb}{{#1}}}}
\newcommand*{\third}[1]{{\setlength{\fboxsep}{0pt}\colorbox{third_rgb}{{#1}}}}

\newcommand{\ie}{\textit{i.e.}\xspace}
\newcommand{\eg}{\textit{e.g.}\xspace}
\newcommand{\Ie}{\textit{I.e.}\xspace}
\newcommand{\Eg}{\textit{E.g.}\xspace}
\newcommand{\etal}{\textit{et al.}\xspace}
\newcommand{\cf}{\textit{cf.}\xspace}
\newcommand{\etc}{\textit{etc.}\xspace}
\newcommand{\wrt}{\textit{w.r.t.}\xspace}
\newcommand{\PAR}[1]{\vskip4pt \noindent{\bf #1~}}

\title{Combining Absolute and Semi-Generalized Relative Poses \\for Visual Localization}

\author[1,2]{Vojtech Panek\orcidlink{0000-0003-0601-7682}}
\author[2]{Torsten Sattler\orcidlink{0000-0001-9760-4553}}
\author[3]{Zuzana Kukelova\orcidlink{0000-0002-1916-8829}}

\affil[1]{Faculty of Electrical Engineering, Czech Technical University (CTU) in Prague}
\affil[2]{Czech Institute of Informatics, Robotics and Cybernetics, CTU in Prague}
\affil[3]{Visual Recognition Group, Faculty of Electrical Engineering, CTU in Prague}

\date{}

\hypersetup{
pdftitle={Combining Absolute and Semi-Generalized Relative Poses for Visual Localization},
pdfsubject={cs.CV},
pdfauthor={Vojtech Panek, Torsten Sattler, Zuzana Kukelova},
pdfkeywords={Visual localization, Camera pose estimation},
}

\begin{document}
\maketitle

\begin{abstract}
    Visual localization is the problem of estimating the camera pose of a given query image within a known scene. Most state-of-the-art localization approaches follow the structure-based paradigm and use 2D-3D matches between pixels in a query image and 3D points in the scene for pose estimation. These approaches assume an accurate 3D model of the scene, which might not always be available, especially if only a few images are available to compute the scene representation. In contrast, structure-less methods rely on 2D-2D matches and do not require any 3D scene model. However, they are also less accurate than structure-based methods. Although one prior work proposed to combine structure-based and structure-less pose estimation strategies, its practical relevance has not been shown. We analyze combining structure-based and structure-less strategies while exploring how to select between poses obtained from 2D-2D and 2D-3D matches, respectively. We show that combining both strategies improves localization performance in multiple practically relevant scenarios.
\end{abstract}

\section{Introduction}
\label{sec:intro}
Estimating the position and orientation from which a given image was taken in a known scene, known as the visual localization task, is an important part of applications such as Augmented Reality~\cite{Lynen2015GetOO, Middelberg2014Scalable6L}, and robotics~\cite{Heng2018ProjectAL,Lim2012RealtimeI6}.

Most state-of-the-art methods establish 2D-3D matches between pixels in a query image and 3D points in the scene~\cite{sarlin2019coarse, Panek2022ECCV, Taira2019InLocIV,  Brachmann2020VisualCR, Cavallari2019TPAMI, Brachmann2023AcceleratedCE}. 
The matches are then used for pose estimation, \eg, by applying a P3P solver~\cite{Persson2018ECCV} inside RANSAC~\cite{Fischler1981RandomSC,Lebeda2012BMVC}. 
Given accurate 3D point positions, this approach leads to precise camera pose estimates. 

An alternative to such structure-based approaches are structure-less methods~\cite{Zheng_2015_ICCV,Bhayani2021CalibratedAP,Zhang2006ImageBL,Dong2023LazyVL,Zhou2020ICRA}.
They estimate the pose of a query image from a set of 2D-2D matches between the query and multiple database images with known poses and intrinsics, \eg, by semi-generalized relative pose estimation~\cite{Zheng_2015_ICCV,Bhayani2021CalibratedAP} or by triangulation~\cite{Zhang2006ImageBL,Dong2023LazyVL,Zhou2020ICRA}. 
Such methods are applicable even without an accurate 3D structure estimate or when there is little overlap between the database images~\cite{Zheng_2015_ICCV}. 
Yet, they are typically less accurate~\cite{Dong2023LazyVL}. 

Rather than using 2D-3D or 2D-2D matches, Camposeco \etal proposed to use both types of matches~\cite{Camposeco2018HybridCP}. 
In each iteration, their proposed Hybrid RANSAC approach first randomly samples a minimal solver. 
Based on the solver, their method draws the required set of 2D-2D and 2D-3D correspondences, estimates the pose, and counts inliers. 
Then it updates the probability distribution that is used to select the solver. 
Their approach does not choose between a structure-based or a structure-less strategy a priori. 
Rather, it selects the appropriate pose estimation strategy per query image based on which approach performs best. 
Experiments with simulated 2D-2D matches showed promising results. 
Yet, we are not aware of visual localization systems that use adaptive strategies such as~\cite{Camposeco2018HybridCP}. 
In this paper, we investigate scenarios in which adaptively choosing between structure-based and structure-less pose estimation strategies improves localization accuracy. 

The contribution of Camposeco \etal is two-fold: (1) they proposed multiple minimal solvers based on both 2D-2D and 2D-3D matches. 
(2) they proposed the Hybrid RANSAC framework, discussed above, that can use multiple solvers with different input sets. 
When selecting the best pose during Hybrid RANSAC, they simply added up the number of inliers for 2D-2D and 2D-3D matches. 
In this paper, we show that such a simple approach for selecting the best camera pose (and thus selecting between different camera pose estimation strategies) does not work well in practice. 
Consequently, we discuss and evaluate multiple equally simple (and thus practically useful) pose selection strategies. 
Besides selecting between different poses inside a (Hybrid) RANSAC framework (as in~\cite{Camposeco2018HybridCP}, we also investigate selecting poses obtained by separate RANSACs for structure-based and structure-less methods. 
Experiments on different real-world datasets show that: (1) the choice of selection strategy has a significant impact on pose accuracy. 
(2) strategies that combine structure-based and structure-less pose estimation approaches are practically relevant.  Especially if only a sparse set of database images is available, or the 3D geometry estimates are inaccurate, they can significantly improve performance. 

In summary, this paper makes the following contribution:
(1) we show that strategies that adaptively select between structure-based and structure-less pose estimation approaches are practically relevant. 
In particular, we show in which scenarios such strategies are useful.
(2) we analyze how to select between these two approaches by selecting an appropriate scoring function for camera pose estimates. 
We show that the choice of the function has a significant impact on the pose accuracy.
(3) we evaluate a simple selection strategy for pairs of pose estimates made using structure-based and structure-less methods.
(4) we will make our code publicly available.

\section{Related Work}
\label{sec:related_work}

\noindent \textbf{Structure-based methods} 
form the leading, well-established branch of visual localization.
They use 2D-3D matches between 2D pixel positions and 3D scene points for camera pose estimation, typically by applying an absolute pose solver~\cite{Persson2018ECCV, Kukelova2016EfficientIO} inside a RANSAC variant~\cite{Fischler1981RandomSC, Raguram2013USACAU, Barth2018MAGSACMS, Barth2019MAGSACAF}.
There are multiple ways how to establish the 2D-3D correspondences.
SfM (Structure from Motion)-based methods~\cite{sarlin2019coarse} start with a set of reference images and build an SfM point cloud.
First, 2D-2D matches are established by extraction of local features~\cite{DeTone2017SuperPointSI, Lowe2004DistinctiveIF, Zhou2020Patch2PixEP, Tyszkiewicz2020DISKLL} with a subsequent matching step~\cite{sarlin2020superglue} or directly with a detector-less matcher~\cite{Sun2021LoFTRDL}.
Image retrieval based on global image descriptors~\cite{Arandjelovi2015NetVLADCA, Jgou2010AggregatingLD, GARL17, RARS19} can be used to pre-filter image pairs to avoid computationally demanding exhaustive matching.
The 2D-2D matches are lifted into 3D by triangulation.
A second way to generate 2D-3D matches is to use explicit or implicit models that capture the geometry and appearance of the scene, such as a mesh~\cite{Panek2022ECCV,Panek2023VisualLU} or a NeRF~\cite{Liu2023NeRFLocVL}.
The triangulation step from the SfM methods is then replaced by the back-projection of keypoints with a rendered depth map.
Another approach are scene coordinate regressors~\cite{Brachmann2017CVPR, Brachmann2018CVPR, Brachmann2020VisualCR, Cavallari20193DV, Cavallari2017CVPR, Cavallari2019TPAMI, Shotton2013SceneCR, Walch2017ICCV, Brachmann2023AcceleratedCE} that learn to predict corresponding 3D points for pixels in a query image. 
The pose accuracy of all structure-based  methods  directly depends on the accuracy of the geometry. 
Pose accuracy tends to degrade if the 3D structure is estimated from a few database images.

\noindent \textbf{Structure-less approaches}
are based on estimating the relative pose to multiple database images (with known absolute poses) solely from 2D-2D matches. 
Since the relative pose is estimated \wrt to multiple database images, the scale of the translation can be recovered. 
Structure-less approaches are typically not used in practice because their accuracy usually does not reach the structure-based methods~\cite{Dong2023LazyVL}.
As the relative pose solvers~\cite{Zheng_2015_ICCV, Larsson2017EfficientSF, Stewnius2006RecentDO, Kukelova2008AutomaticGO} generally use more matches than the absolute pose solvers, they also need more RANSAC iterations.
The main advantage (which we actively exploit in this paper) is their independence from any scene geometry model.
We show that structure-less approaches more gracefully handle representations with less accurate scene geometry, thus creating a strong motivation for using such approaches in practice.

\noindent \textbf{Adaptive approaches} 
use both 2D-2D and 2D-3D matches for camera pose estimation.
These methods are not deeply studied in the literature, and we are only aware of one such approach~\cite{Camposeco2018HybridCP} (described in Sec.~\ref{sec:method}). 
While~\cite{Camposeco2018HybridCP} showed promising results, they did not evaluate their approach in a practical setting. 
This paper closes this gap, showing that adaptively choosing whether to use structure-based or structure-less pose estimation strategies can improve performance. 
We build upon~\cite{Camposeco2018HybridCP} but focus on making  adaptive approaches work in practice. 
In particular, we show the performance of adaptive approaches heavily depends on the scoring function used to choose between poses and that the scoring function used in~\cite{Camposeco2018HybridCP} is not a good choice. 
We show that selecting between poses after RANSAC, rather than within each RANSAC iteration, can boost performance in certain scenarios. 
In general, we showcase different practical scenarios where adaptive approaches excel.

\section{Adaptive Pose Selection}
\label{sec:method}
This paper aims to understand whether combining structure-based and structure-less pose estimation approaches can improve performance in practice. 
To answer this question, we design a simple approach that adaptively chooses between the two strategies depending on which one leads to better pose estimates. 

\PAR{A simple adaptive pose estimation strategy.} 
Given sets of 2D-2D and 2D-3D matches, we employ two pose solvers inside a RANSAC loop: 
a P3P solver~\cite{Persson2018ECCV} that uses 2D-3D matches and the E5+1 solver~\cite{Zheng_2015_ICCV} that uses 2D-2D matches. 
In each iteration, we draw two samples, one from the 2D-3D matches for the P3P solver and one from the 2D-2D matches for the E5+1 solver. 
The poses predicted by both solvers are then evaluated on both 2D-2D and 2D-3D matches, and the current best pose estimate is updated if a new best pose is found. 
For 2D-3D matches, we measure the reprojection error. 
For 2D-2D matches, we measure the Sampson error~\cite{Sampson1982FittingCS, Hartley2001MultipleVG}.
Each newly found best pose is refined using local optimization~\cite{Lebeda2012BMVC}. 

Our approach is a simplified version of~\cite{Camposeco2018HybridCP}, where we focus on only two solvers\footnote{In contrast to many of the other solvers used in~\cite{Camposeco2018HybridCP}, both P3P and E5+1 have very efficient, and equally importantly publicly available, implementations. 
In contrast to the other solvers from~\cite{Camposeco2018HybridCP}, both are easy to implement. 
Out of all solvers used in~\cite{Camposeco2018HybridCP}, the P3P and E5+1 solvers thus seem to be the most practically relevant.} and apply each solver in each iteration. 
This simplification is a deliberate choice. 
Combining 2D-3D and 2D-2D inliers that are evaluated using different errors is, in general, not straightforward. 
Thus, the main question we pose here is the following: how to define the best pose and how to perform local optimization using these two sources of data? 
Our simplified approach is sufficient to answer this question. 
However, the strategies for pose selection and optimization described below can directly be applied in the Hybrid RANSAC scheme from~\cite{Camposeco2018HybridCP} if efficiency is of  concern.

\PAR{Camera pose scoring.} 
Given a single type of matches, \eg, only 2D-3D matches, the common strategy is to rate pose estimates based on the number $I$ of inliers or based on minimizing a robust cost function. 
An example of the latter is the MSAC scoring function $M = \sum_i \min(e_i^2, t^2)$~\cite{Torr2000MLESACAN}, where $e_i$ is the error of the $i$-th match \wrt the current pose estimate and $t$ is the threshold used in RANSAC to distinguish between inliers and outliers.  

For two types of matches, we obtain the numbers of inliers $I_\text{3D}$ and $I_\text{2D}$ for 2D-3D matches and 2D-2D matches, respectively. 
Similarly, we obtain MSAC scores $M_\text{3D}$ and $M_\text{2D}$ for 2D-3D and 2D-2D matches, respectively. 
As shown in Sec.~\ref{sec:experimental_evaluation}, how we combine these scores to select the best pose found so far inside RANSAC has a significant impact on the overall pose accuracy.
As such, whether our approach is able to adapt to the quality of the two sets of matches strongly depends on the pose scoring function.

The hybrid RANSAC approach from~\cite{Camposeco2018HybridCP} simply \textit{sums up the numbers of inliers} (denoted as \textit{sum. inl.} in the following) to obtain a single score 
\begin{equation}
    s_\text{sum inl.} = I_\text{3D} + I_\text{2D} \enspace . \label{eq:sum_inliers}
\end{equation}
Typically, there are more 2D-2D than 2D-3D matches\footnote{For example, the state-of-the-art structure-based HLoc~\cite{sarlin2019coarse,sarlin2020superglue} method first performs 2D-2D matching between a query and similar database images. It then uses an SfM model to determine which of the matching 2D pixels in the database images has a corresponding 3D point in the SfM model. Thus, not every 2D-2D match necessarily results in a 2D-3D match.}. 
As such, $s_\text{sum inl.}$ might give too much weight to 2D-2D matches. 
Potential ways to avoid this problem are to
\textit{sum up the inlier ratios} (\textit{sum. inl. rat.}) 
\begin{equation}
    s_\text{sum inl. ratios} = I_\text{3D} / N_\text{3D} + I_\text{2D} / N_\text{2D} \enspace , \label{eq:sum_inlier_ratios}
\end{equation}
where $N_\text{3D}$ and $N_\text{2D}$ are the numbers of 2D-3D and 2D-2D matches, or to \textit{multiply the numbers of inliers} (\textit{mult. inl.})
\begin{equation}
    s_\text{mult inl.} = I_\text{3D} \cdot I_\text{2D} \enspace . \label{eq:mult_inlier}
\end{equation}
Note that multiplying the number of inliers or inlier ratios leads to the same scoring function as they both differ by a constant term ($1/\left(N_\text{3D}\cdot N_\text{2D}\right)$).
Similarly, we can either \textit{sum up the individual MSAC scores} (\textit{sum. msac})
\begin{equation}
    s_\text{sum msac} = M_\text{3D} / t_\text{3D}^2 + M_\text{2D} / t_\text{2D}^2 \enspace , \label{eq:sum_msac}
\end{equation}
or \textit{multiply the two MSAC scores} (\textit{mult. msac}),
\begin{equation}
    s_\text{mult msac} = M_\text{3D} \cdot M_\text{2D} \enspace . \label{eq:mult_msac}
\end{equation}

Note that in Eq.~\ref{eq:sum_msac}, we normalize the MSAC scores by dividing them by the inlier threshold $t_\text{3D}$ used for the reprojection error, respectively, the threshold $t_\text{2D}$ used for the Sampson error, to take into account that both have different magnitudes.
Multiplication of the two values emphasizes poses that are good for both modalities (2D-2D and 2D-3D), similar to the logical ”and” operator.
\Eg, a pose with 50 2D-3D and 50 2D-2D inliers is preferred over a pose with 10 2D-3D and 100 2D-2D inliers. 
Also note that the scores based on the numbers of inliers or inlier ratios are maximized, while scores based on MSAC scores are minimized.

The scoring functions discussed above aim to select a pose that is consistent with both sets of matches. 
In some scenarios, the set of 2D-3D correspondences can be unreliable, \eg, when using very noisy depth maps to obtain 2D-3D matches from a set of 2D-2D matches and finding a pose that is consistent with both sets of correspondences is not meaningful.
For such cases, we consider a strategy, termed \texttt{Select}, that selects between the best pose estimated from the 2D-2D and the best pose estimated from the 2D-3D matches. 
For this selection, as the 2D-3D matches might be unreliable, we rely on the 2D-2D matches alone.\footnote{2D-3D matches are obtained by "lifting" 2D-2D matches, \ie, by associating matching 2D features with their corresponding 3D points. Even if the 3D points are inaccurate, and the 2D-3D matches thus are unreliable, it seems reasonable to assume that the 2D-2D matches are reliable.} 
Given that P3P poses, in our experience, are more reliable than E5+1 poses when estimated from accurate 2D-3D matches, we thus prefer to select P3P poses. 
Let $I_\text{2D}^\text{P3P}$ and $I_\text{2D}^\text{E5+1}$ be the number of 2D-2D match inliers found by P3P respectively E5+1.
We select the P3P pose if 
\begin{equation}
    I_\text{2D}^\text{P3P} > \alpha \cdot I_\text{2D}^\text{E5+1}
\end{equation}
for a threshold $\alpha > 0$ and E5+1 pose otherwise.
Intuitively, we select the P3P pose if it is consistent with the 2D-2D inliers of the E5+1 pose, assuming that in this case, the 2D-3D matches have to be accurate and reliable. 
Note that the \texttt{Select} approach does not need to be used inside RANSAC. 
It is sufficient to run both P3P and E5+1 RANSAC in parallel and to select the more appropriate pose using the \texttt{Select} strategy afterwards.

\PAR{Local Optimization.} 
The goal of local optimization in RANSAC is to reduce the impact of measurement noise on pose accuracy~\cite{Lebeda2012BMVC}.
Hybrid RANSAC~\cite{Camposeco2018HybridCP} optimizes poses only based on 2D-3D matches.
We investigate two ways how to locally optimize the pose hypotheses.
The first option is to use a \textit{hybrid} refinement, optimizing over both 2D-2D and 2D-3D matches.
The second option (denoted as \textit{split}) is to optimize only one of the scores, selected based on the pose solver, which generated the pose hypothesis that is being optimized.
If the P3P solver generates the pose, the 2D-3D error is optimized. 
If the E5+1 solver generates the pose, the 2D-2D error is optimized. 
The intuition is that in the case of imprecise scene geometry, the E5+1 solver will generate the final pose estimate, which will not be skewed by the inaccurate 2D-3D matches during the local optimization.
In both cases, local optimization is implemented as a non-linear refinement of the reprojection and Sampson error, respectively, using the implementation provided in PoseLib~\cite{PoseLib}. 

\PAR{A simple hybrid localization approach.} 
We integrate the simplified pose estimation approach into a standard visual localization pipeline~\cite{sarlin2019coarse,sarlin2020superglue,Panek2022ECCV}: 
given a set of database images with known poses and intrinsics, we compute a geometric representation of the scene from these images in an offline step.
For a given query image, we use image retrieval to identify a set of potentially relevant database images and match features between the query and each retrieved database image. 
For each of the resulting 2D-2D matches, we use the geometric representation to identify whether there is a 3D point in the scene corresponding to the 2D position in the database image. 
This look-up, which can, for example, be based on a sparse SfM point cloud of the scene~\cite{Irschara2009CVPR,Sattler2012BMVC,sarlin2019coarse} or based on dense depth maps~\cite{Panek2022ECCV}, results in a set of 2D-3D matches. 
All the 2D-3D and 2D-2D matches are then used for pose estimation.

\section{Experimental evaluation}
\begin{figure*}[t!]
    \centering
    \includegraphics[width=0.96\textwidth,trim={0.0cm 0 0.0cm 0},clip]{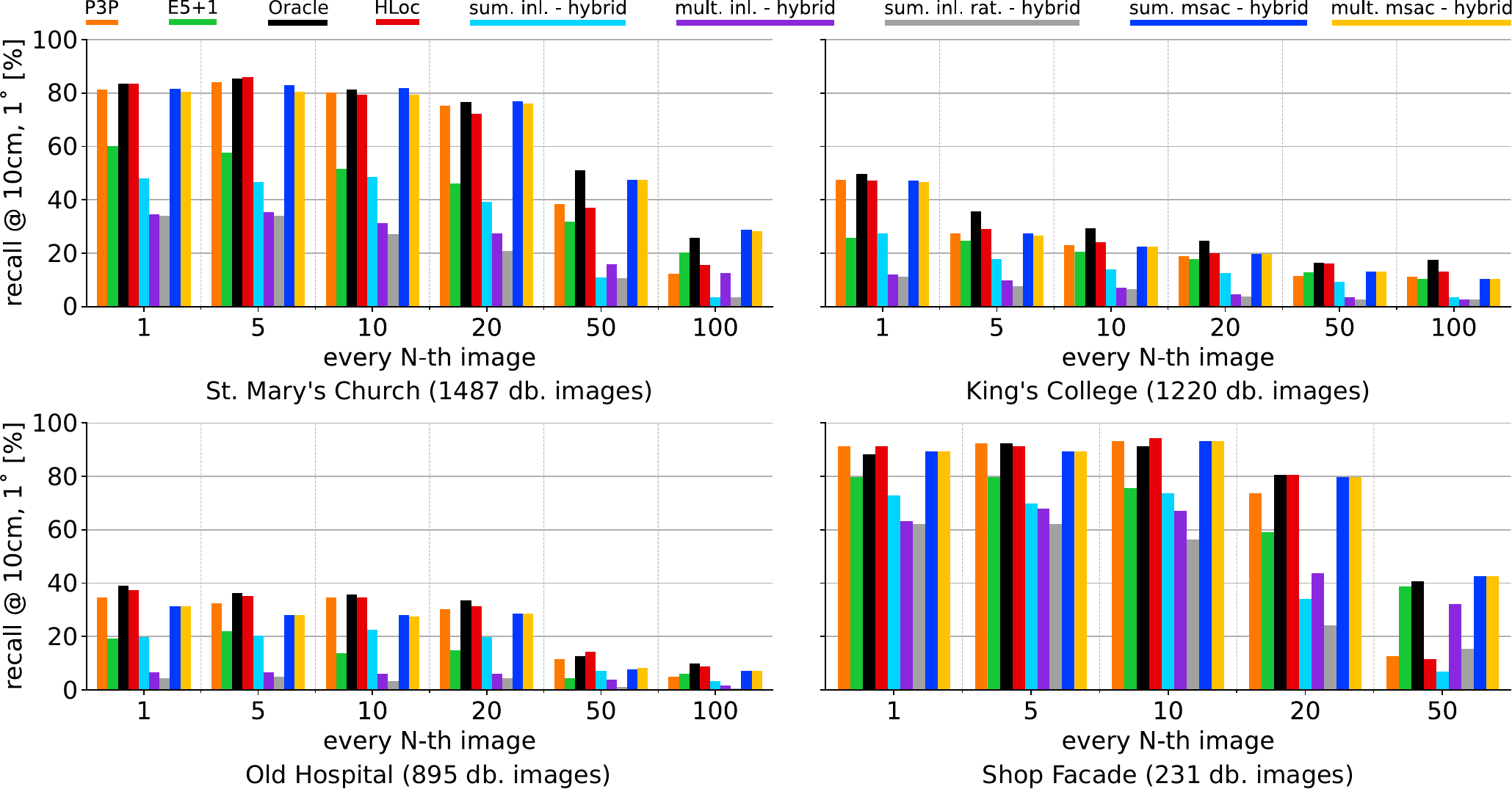}
    \caption{Ablating  scoring functions on the Cambridge Landmarks dataset~\cite{Kendall2015PoseNetAC}. We report the percentage of images localized within 10 cm and 1 degree of the ground truth. The scene is represented using SfM point clouds computed using every N-th database image.}
    \label{fig:score_cambridge_real}
\end{figure*}
\label{sec:experimental_evaluation}
The focus of this paper is to answer the question of whether visual localization approaches that use both 2D-2D and 2D-3D matches are useful in practice. 
To this end, we pose several important practical questions: 
(A) Is traditional structure-based localization always outperforming structure-less localization in terms of precision? 
(B) Is it possible to design a simple and effective method that combines these two approaches, which, without additional knowledge of the scene geometry, takes advantage of both? 
(C) Are these two individual localization strategies performing differently for different types of scene representation? 
(D) In which use cases do adaptive approaches offer improved performance?
To answer these questions, we evaluate different variants of the proposed adaptive scheme and compare them with a structure-based and a structure-less baseline in different settings.

\subsection{Experimental setup}
\noindent \textbf{Datasets.} We evaluate on multiple publicly available benchmarks. 
The 7 Scenes dataset~\cite{Glocker2013RealtimeRC, Shotton2013SceneCR} (in Sec.~\ref{sec:supp_mat}) captures small room-scale scenes, while the Gangnam Station dataset~\cite{Lee2021LargescaleLD} shows a large-scale indoor public facility. 
Database and query images were captured at (roughly) the same time for Cambridge Landmarks, 7 Scenes, and Gangnam Station, and thus depict the scenes under very similar conditions.
In contrast, the Aachen~\cite{Zhang2020ARXIV, Sattler2018CVPR, Sattler2012BMVC} and Extended CMU Seasons~\cite{Sattler2018CVPR, Badino2011} datasets contain query images taken under different conditions compared to the database images, \eg, day-night and seasonal changes, making them significantly more challenging.

\begin{table*}
\setlength{\tabcolsep}{3pt}
\begin{center}
\small{
\begin{tabular}{l l l l l l l l l}
 Method & Scoring & LO & 
 1/1 & 1/5 & 1/10 & 1/20 & 1/50 \\
 \cline{4-8}\noalign{\smallskip}
\texttt{P3P} & & & \second{77.0} / 86.9 / \first{99.0} & 67.5 / 83.8 / \first{93.7}& 57.1 / \third{73.8} / 83.8 & 30.9 / 50.3 / 67.5 & 7.3 / 14.7 / 33.5 \\
\texttt{E5+1} & & & 69.1 / 86.4 / \third{96.3} & 62.3 / 75.4 / 90.1 & 48.2 / 63.9 / 81.2 & 27.2 / 45.0 / 66.5 & \second{15.7} / 25.1 / 39.8 \\ \hline
\multirow{10}{*}{\texttt{Adaptive}} & \multirow{2}{*}{sum. inl.} & hybrid & 52.9 / 72.3 / 91.1 & 26.2 / 39.3 / 71.2 & 19.4 / 30.4 / 59.2 & 8.4 / 11.5 / 34.0 & 1.6 / 4.2 / 12.6 \\
 &  & split & 52.4 / 70.7 / 91.6 & 24.1 / 39.8 / 71.2 & 14.7 / 27.7 / 59.2 & 8.9 / 11.0 / 34.6 & 1.0 / 2.6 / 12.6 \\ \cline{2-8}
&  \multirow{2}{*}{mult. inl.} & hybrid & 59.2 / 81.2 / \second{97.9} & 48.7 / 69.6 / \second{93.2}& 35.6 / 61.8 / 82.7 & 23.0 / 41.4 / 68.6 & 8.4 / 15.2 / 34.6 \\ 
&  & split & 59.7 / 80.6 / \second{97.9} & 47.6 / 69.1 / 91.6 & 35.6 / 61.8 / 82.7 & 19.9 / 38.2 / 69.1 & 5.8 / 14.1 / 35.6 \\ \cline{2-8}
& \multirow{2}{*}{sum. inl. rat.} & hybrid & 55.5 / 80.1 / \second{97.9} & 44.5 / 71.2 / \first{93.7}& 30.9 / 59.2 / \third{86.4} & 16.8 / 40.8 / 71.2 & 5.8 / 13.6 / 42.9 \\
&  & split & 55.5 / 80.1 / \second{97.9} & 44.5 / 71.2 / \first{93.7}& 30.9 / 59.2 / \third{86.4} & 16.8 / 40.8 / 71.2 & 5.8 / 13.6 / 42.9 \\ \cline{2-8}
& \multirow{2}{*}{sum. msac} & hybrid & \first{78.5} / \second{89.0} / \first{99.0} & \third{68.6} / 83.8 / \first{93.7} & 58.6 / 73.8 / \second{86.9} & \second{39.3} / \third{56.5} / 71.7 & \first{16.2} / \third{28.8} / \second{48.2} \\
&  & split & \third{75.9} / \third{88.0} / \first{99.0} & 68.1 / 82.7 / \first{93.7} & 58.6 / 73.8 / \second{86.9} & 35.1 / 52.4 / 71.2 & 13.1 / 27.2 / 46.1 \\ \cline{2-8}
& \multirow{2}{*}{mult. msac} & hybrid & \first{78.5} / \second{89.0} / \first{99.0} & \second{69.1} / 83.8 / \first{93.7}& \third{59.2} / \second{74.9} / \first{87.4} & \first{40.8} / \first{57.6} / 73.8 & \second{15.7} / 27.7 / 44.5 \\
 &  & split & 75.4 / \third{88.0} / \first{99.0} & 67.5 / \third{84.3} / \first{93.7}& \second{60.2} / \first{75.4} / \second{86.9} & \third{36.1} / 56.0 / 72.3 & \third{13.6} / 24.1 / 45.0 \\  \hline
\texttt{Select} $\alpha=0.7$ &  &  & 73.3 / 86.4 / \second{97.9} & 65.4 / \second{84.8} / \second{93.2} & 51.3 / \second{74.9} / \second{86.9} & 33.5 / \first{57.6} / \first{75.9} & 11.5 / \third{28.8} / \first{49.2} \\ 
\texttt{Select} $\alpha=0.8$ &  &  & 73.3 / 86.9 / \second{97.9} & 65.4 / \second{84.8} / \second{93.2} & 51.3 / \second{74.9} / \second{86.9} & 33.5 / \third{56.5} / \second{74.9} & 12.0 / \second{29.3} / \first{49.2} \\ 
\texttt{Select} $\alpha=0.9$ &  &  & 73.3 / 86.4 / \second{97.9} & 64.9 / \third{84.3} / \third{92.7} & 51.3 / \third{74.3} / \third{86.4} & 33.5 / \second{57.1} / \third{74.3} & 12.6 / \first{30.4} / \first{49.2} \\ 
\texttt{Select} $\alpha=1.0$ &  &  & 57.6 / 78.0 / 96.9 & 56.5 / 79.6 / 90.1 & 48.2 / 68.6 / 82.7 & 29.8 / 52.9 / 71.7 & \third{13.6} / 26.7 / \third{47.6} \\ \hline
\texttt{HLoc}~\cite{sarlin2019coarse,sarlin2020superglue} & & & \second{77.0} / \first{89.5} / \first{99.0} & \first{70.7} / \first{85.9} / \second{93.2}& \first{60.7} / \first{75.4} / 84.3 & 35.1 / 54.5 / 66.5 & 8.4 / 17.3 / 31.9 \\
\end{tabular}
}
\end{center}
\caption{Ablating scoring functions and local optimization (LO) strategies on the nighttime queries of the Aachen Day-Night v1.1 dataset~\cite{Zhang2020ARXIV, Sattler2018CVPR, Sattler2012BMVC} (6,697 database images). Showing localization recalls (higher is better) at the pose error thresholds of (0.25m, 2°) / (0.5m, 5°) / (5m, 10°).
We mark the \first{best}, \second{second-}, and \third{third-best} results per column.
}
\label{tab:score_aachen}
\end{table*}

\begin{table*}
\setlength{\tabcolsep}{3pt}
\begin{center}
\small{
\begin{tabular}{l l l l l l l l l}
 & & & & \multicolumn{2}{l}{day} && \multicolumn{2}{l}{night} \\
 \cline{5-6} \cline{8-9}\noalign{\smallskip}
 & RE thr. & in. thr. & rel. thr. & Metric3D & AC13 mesh && Metric3D & AC13 mesh \\
 \cline{2-9}\noalign{\smallskip}
\texttt{Adaptive} - mult. msac - hybrid & - & - & - & 52.9 / 78.8 / \second{97.9} & \first{87.3} / \first{94.7} / \first{98.5} && 42.9 / 71.7 / \second{97.4} & \second{72.3} / \second{88.0} / \second{97.9} \\
\texttt{Adaptive} - mult. msac - hybrid & 4.0 & 3/20 & - & 52.5 / 71.7 / 94.3 & 83.4 / 91.5 / 96.8 && 37.7 / 61.3 / 92.7 & 65.4 / \third{87.4} / 94.8 \\
\texttt{Adaptive} - mult. msac - hybrid & 4.0 & 3/20 & 0.8 & 68.8 / 81.3 / 96.0 & \third{84.1} / 92.5 / 97.2 && 53.4 / 72.8 / 93.7 & 67.0 / 86.4 / 96.3 \\
\texttt{Adaptive} - mult. msac - hybrid & 4.0 & 18/20 & - & 70.6 / 84.1 / 96.8 & 72.6 / 86.5 / 95.5 && 58.1 / 75.9 / \second{97.4} & 57.6 / 74.9 / 93.2 \\
\texttt{Adaptive} - mult. msac - hybrid & 2.0 & 3/20 & - & 55.2 / 72.5 / 94.5 & 82.9 / 91.6 / 96.8 && 40.8 / 58.1 / 91.1 & 64.9 / 86.9 / 94.8 \\
\texttt{Adaptive} - mult. msac - hybrid & 2.0 & 3/20 & 0.8 & 69.7 / 82.8 / 94.9 & 83.3 / 92.4 / 97.1 && 51.8 / 71.7 / 93.7 & \third{68.6} / 85.9 / 95.3 \\
\texttt{Adaptive} - mult. msac - hybrid & 2.0 & 18/20 & - & 69.7 / 83.7 / 97.0 & 66.1 / 82.4 / 94.2 && 60.2 / 77.5 / 95.8 & 57.1 / 74.3 / 91.1 \\
\texttt{Adaptive} - sum. msac - hybrid & - & - & - & 51.6 / 77.7 / \first{98.2} & \second{86.3} / \third{93.8} /\first{98.5} && 41.9 / 67.0 / \second{97.4} & \first{73.3} / \first{88.5} / \second{97.9} \\
\texttt{Adaptive} - sum. msac - hybrid & 4.0 & 3/20 & - & 60.7 / 81.6 / 97.6 & 83.7 / 91.7 / \third{97.9} && 48.7 / 73.8 / \third{96.9} & 66.0 / \second{88.0} / \second{97.9} \\
\texttt{Adaptive} - sum. msac - hybrid & 4.0 & 3/20 & 0.8 & 75.5 / 89.4 / 97.7 & 83.4 / 92.2 / \second{98.1} && \third{64.4} / 81.2 / \first{97.9} & 66.0 / 86.4 / \second{97.9} \\
\texttt{Adaptive} - sum. msac - hybrid & 4.0 & 18/20 & - & 78.3 / 90.3 / \third{97.8} & 76.8 / 90.0 / 97.8 && 62.8 / \third{83.2} / \first{97.9} & 64.4 / 81.7 / \second{97.9} \\
\texttt{Adaptive} - sum. msac - hybrid & 2.0 & 3/20 & - & 65.5 / 84.5 / 97.8 & 83.3 / 91.6 / \third{97.9} && 52.9 / 75.9 / \third{96.9} & 66.5 / \first{88.5} / \third{97.4} \\
\texttt{Adaptive} - sum. msac - hybrid & 2.0 & 3/20 & 0.8 & \first{79.0} / \first{91.0} / \third{97.8} & 82.3 / 91.7 / \second{98.1} && 63.4 / \third{83.2} / \first{97.9} & 67.0 / 86.4 / \second{97.9} \\
\texttt{Adaptive} - sum. msac - hybrid & 2.0 & 18/20 & - & \second{78.9} / \second{90.9} / 97.7 & 77.7 / 90.8 / 97.7 && \second{64.9} / \second{84.8} / \first{97.9} & 64.4 / 83.8 / \second{97.9} \\
\texttt{P3P} & - & - & - & 19.9 / 39.3 / 93.2 & \first{87.3} / \second{94.2} / \first{98.5} && 18.8 / 46.6 / 96.3 & \second{72.3} / \first{88.5} / \first{98.4} \\
\texttt{E5+1} & - & - & - & \third{78.5} / \third{90.7} / \third{97.8} & 78.5 / 90.7 / 97.8 && \first{66.0} / \first{85.3} / \first{97.9} & 66.0 / 85.3 / \second{97.9}
\end{tabular}
}
\end{center}
\caption{Experiments using 3D geometry from a monocular depth estimator or an MVS mesh of the Aachen Day-Night v1.1 dataset~\cite{Zhang2020ARXIV, Sattler2018CVPR, Sattler2012BMVC}. Note that \texttt{E5+1} is independent on the geometry. Showing localization recalls (higher is better) at the pose error thresholds of (0.25m, 2°) / (0.5m, 5°) / (5m, 10°).}
\label{tab:aachen_depth_exps}
\end{table*}

\noindent \textbf{Methods.} We use the following set of methods for our experimental evaluation: 
(1) \texttt{P3P} computes the camera pose from  2D-3D correspondences by applying a P3P solver~\cite{Persson2018ECCV} within RANSAC with local optimization; 
(2) \texttt{E5+1} computes the camera pose from 2D-2D matches by applying the E5+1 solver ~\cite{Zheng_2015_ICCV} within RANSAC with local optimization; 
(3) \texttt{"Oracle"} compares the two poses generated by the \texttt{P3P} and \texttt{E5+1} baselines and selects the one that has a smaller absolute error~\footnote{The absolute error is defined as the maximum of position error in centimeters and orientation error in degrees} compared to ground truth pose~\footnote{Note that \texttt{"Oracle"} does not provide an absolute upper bound, as it selects only from the two poses estimated by \texttt{P3P} and \texttt{E5+1}. \Ie, the selected pose was computed from either only 2D-2D or only 2D-3D matches. Better poses can potentially be obtained by taking both sets of matches into account.
Still, \texttt{"Oracle"} has access to the ground truth camera poses, which are not available to any of the other methods.};
(4) \texttt{HLoc}~\cite{sarlin2019coarse, sarlin2020superglue} is a state-of-the-art structure-based localization approach\footnote{The difference in comparison to the \texttt{P3P} baseline is in implementation. \texttt{P3P} is based on the PoseLib~\cite{PoseLib} framework. \texttt{HLoc} is build on top of COLMAP~\cite{schoenberger2016sfm, schoenberger2016mvs, schoenberger2016vote} and implements covisibility clustering~\cite{Sarlin2018LeveragingDV}.};
(5) \texttt{Adaptive} is the approach described in Sec.~\ref{sec:method} that selects poses inside RANSAC based on 2D-2D and 2D-3D matches. 
\texttt{Select} is the approach described in Sec.~\ref{sec:method} that selects either the \texttt{P3P} or \texttt{E5+1} pose based on the number of 2D-2D inliers.
Except \texttt{HLoc}, all other baselines are implemented in PoseLib~\cite{PoseLib} (\texttt{P3P}, \texttt{E5+1}) or based on PoseLib (\texttt{Adaptive}, \texttt{"Oracle"}, \texttt{Select}).

\noindent \textbf{Evaluation metric.}
We follow the literature and report the percentage of query images localized within certain error bounds, \eg, the percentage of images with a position error below 5 cm and a rotation error below 5 degrees~\cite{Shotton2013SceneCR}.
We also report median position and orientation errors for the 7 Scenes datasets (in Sec.~\ref{sec:supp_mat}).

\begin{table*}
\setlength{\tabcolsep}{3pt}
\begin{center}
\small{
\begin{tabular}{l l l l l l l}
 & urban &  & suburban &  & park &  \\
 \cline{2-7}\noalign{\smallskip}
 & 1/1 & 1/11 & 1/1 & 1/11 & 1/1 & 1/11 \\
 \cline{2-7}\noalign{\smallskip}
\texttt{P3P}
& \first{95.7} / \first{99.0} / \first{99.6} & 63.6 / 75.3 / 89.6 &  \first{95.0} / \first{98.0} / \first{99.4} & 50.3 / 69.3 / 90.9 & \first{90.7} / \first{94.4} / \first{96.3} & 30.8 / 45.3 / 69.1 \\
\texttt{E5+1}
& 71.9 / 87.2 / 96.9 & 20.9 / 35.8 / 74.2 & 70.4 / 85.6 / 94.6 & 13.0 / 25.7 / 65.9 & 54.6 / 68.0 / 78.5 & 4.9 / 10.8 / 38.3 \\
\texttt{Adaptive}
& \first{95.7} / 98.9 / \first{99.6} & \first{70.5} / \first{81.1} / \first{91.5} & \first{95.0} / 97.8 / 98.9 & \first{58.0} / \first{76.2} / \first{92.8} & 89.5 / 93.2 / 94.5 & \first{37.6} / \first{51.9} / \first{71.2} \\
\end{tabular} \vspace{-8pt}
}
\end{center}
\caption{Results on the  Extended CMU Seasons dataset~\cite{Sattler2018CVPR, Badino2011}. We report the percentage of query images localized within (0.25m, 2°) / (0.5m, 5°) / (5m, 10°) of the ground truth poses when using every database image (1/1) respectively every 11th database image (1/11) to build a SfM model of the scene. For sparse representations, the \texttt{Adaptive} approach, using \textit{mult. msac} scoring with \textit{hybrid} local optimization, clearly performs best, while offering competitive performance when using all images.}
\label{tab:score_cmuse_static}
\end{table*}

\begin{table*}[t!]
\setlength{\tabcolsep}{6pt}
\begin{center}
\small{
\begin{tabular}{l l l l l l l}
 & & \multicolumn{2}{l}{day} && \multicolumn{2}{l}{night} \\
 \cline{3-4} \cline{6-7}\noalign{\smallskip}
 & $\alpha$ & Metric3D & AC13 mesh && Metric3D & AC13 mesh \\
 \cline{2-7}\noalign{\smallskip}
\texttt{Select} & 0.7  & 34.8 / 56.1 / 96.1 & \first{87.3} / \second{94.2} / \first{98.5} && 29.8 / 58.1 / \third{96.9} & \first{72.3} / \first{88.5} / \first{98.4} \\
\texttt{Select} & 0.8  & 41.3 / 62.9 / 97.2 & \first{87.3} / \second{94.2} / \first{98.5} && 34.0 / 63.9 / \second{97.4} & \second{71.7} / \second{88.0} / \second{97.9} \\
\texttt{Select} & 0.9  & 49.4 / 70.3 / \third{97.3} & \first{87.3} / \first{94.3} / \first{98.5} && 40.3 / 70.7 / \second{97.4} & \second{71.7} / \second{88.0} / \second{97.9} \\
\texttt{Select} & 0.95 & 54.4 / 75.5 / \second{97.5} & \second{87.0} / \third{94.1} / \second{98.4} && 47.6 / 78.5 / \second{97.4} & \second{71.7} / \second{88.0} / \second{97.9} \\
\texttt{Select} & 1.0  & \third{76.9} / \third{89.9} / \first{97.8} & \third{84.6} / 93.2 / 97.9 && \second{64.9} / \first{86.4} / \first{97.9} & \third{69.1} / 85.9 / \second{97.9} \\
\texttt{P3P} & -       & 19.9 / 39.3 / 93.2 & \first{87.3} / \second{94.2} / \first{98.5} && 18.8 / 46.6 / 96.3 & \first{72.3} / \first{88.5} / \first{98.4} \\
\texttt{E5+1} & -      & \second{78.5} / \second{90.7} / \first{97.8} & 78.5 / 90.7 / 97.8 && \first{66.0} / \second{85.3} / \first{97.9} & 66.0 / 85.3 / \second{97.9} \\
\texttt{Adaptive} - sum. msac - hybrid & - & \first{79.0} / \first{91.0} / \first{97.8} & 82.3 / 91.7 / \third{98.1} && \third{63.4} / \third{83.2} / \first{97.9} & 67.0 / \third{86.4} / \second{97.9} \\
(2.0 px, 3/20, 80\%) & & & && & 
\end{tabular}
}
\end{center}
\caption{Experiments with \texttt{Select} method using 3D geometry from a monocular depth estimator or an MVS mesh of the Aachen Day-Night v1.1 dataset~\cite{Zhang2020ARXIV, Sattler2018CVPR, Sattler2012BMVC}. Note that \texttt{E5+1} is independent on the geometry. Showing localization recalls (higher is better) at the pose error thresholds of (0.25m, 2°) / (0.5m, 5°) / (5m, 10°).}
\label{tab:select}
\end{table*}

\subsection{Ablation study}
We first evaluate the impact of scoring functions and local optimization strategies on pose accuracy. 

\noindent \textbf{Ablating scoring functions.} Fig.~\ref{fig:score_cambridge_real} shows the percentage of query images localized within 10 cm and 1 degree of their ground truth poses for the \texttt{Adaptive} approach when using different scoring functions using the \textit{hybrid} local optimization strategy. 
In addition, we report results obtained for the \texttt{P3P}, \texttt{E5+1}, \texttt{"Oracle"}, and \texttt{HLoc} baselines. 
Methods that use 2D-3D matches represent the scene as a Structure-from-Motion (SfM) point cloud. 
2D-2D matches are obtained by matching the SuperPoint~\cite{DeTone2017SuperPointSI} features with the SuperGlue~\cite{sarlin2020superglue} matcher between the query image and visually similar database images found via NetVLAD-based~\cite{Arandjelovi2015NetVLADCA} image retrieval. 
The resulting 2D-2D matches are then lifted to 2D-3D matches via the SfM model~\cite{sarlin2019coarse}. 
We report results for scene representations obtained using every $N$-th image from the database image sequences provided by the dataset, for varying $N$. 
Increasing $N$ leads to fewer database images being used. 
This increases the distance between database images, making the matching between database images harder. 
Thus, 3D points are triangulated from fewer images (potentially decreasing their accuracy), and fewer 3D points are generated overall. 
Fig.~\ref{fig:score_cambridge_real} shows results for the Cambridge Landmarks dataset. 
Tab.~\ref{tab:score_aachen} reports results for the Aachen Day-Night v1.1 dataset~\cite{Zhang2020ARXIV, Sattler2018CVPR, Sattler2012BMVC}.

The results shown in Fig.~\ref{fig:score_cambridge_real} lead to multiple interesting observations: 
(\textbf{1}) For densely sampled scenes, \ie, small values of $N$, structure-based localization (\texttt{P3P}) significantly outperforms structure-less localization (\texttt{E5+1}). 
As sparsity increases, \ie, as $N$ increases, the performance of \texttt{P3P} degrades faster than that of \texttt{E5+1}. 
For sparsely represented scenes, \eg, the Shop Facade scene for $N=50$\footnote{For reference, the Shop Facade scene contains a total of only 231 database images.} or Aachen for $N=50$, \texttt{E5+1} can (significantly) outperform \texttt{P3P}, as well as the \texttt{HLoc} baseline.  
Thus, the answer to \textit{question (A)} is that \textit{structure-less approaches can outperform structure-based approaches}.  
(\textbf{2}) the choice of the scoring function has a significant impact on pose accuracy. 
Simply adding the numbers of 2D-2D and 2D-3D inliers, as done in~\cite{Camposeco2018HybridCP}, performs worse than summing or multiplying MSAC scores. 
MSAC scores are known to select more accurate poses than inlier counting when using the same type of matches~\cite{Lebeda2012BMVC}, \eg, only 2D-2D matches. 
Unsurprisingly, summing or multiplying MSAC scores also provides a better estimate of pose quality when using different types of matches, with both strategies performing similarly. 
(\textbf{3}) the \texttt{Adaptive} approach, using an MSAC-based scoring function, consistently performs similarly or better than both \texttt{P3P} and \texttt{E5+1}. 
In particular, the \texttt{Adaptive} approach achieves a similar accuracy as \texttt{P3P} for small values of $N$ while performing similarly or better to \texttt{E5+1} for large values of $N$. 
In addition, it is competitive with the \texttt{"Oracle"} baseline, which selects the pose from the poses generated by \texttt{P3P} and \texttt{E5+1} that is closest to the ground truth pose, and the state-of-the-art \texttt{HLoc} approach. 
As such, the answer to \textit{question (B)} is that \textit{it is possible to design a simple and effective method that combines the best of structure-based and structure-less pose estimation strategies}. 

\noindent \textbf{Ablating local optimization strategies.} 
Tab.~\ref{tab:score_aachen} evaluates the impact of the type of local optimization strategy on camera pose accuracy. 
Overall, the \textit{hybrid} strategy performs similarly or better than the \textit{split} strategy, especially for higher values of $N$. 
This does not seem surprising, as the \textit{hybrid} approach optimizes over both match types, while the \textit{split} strategy only uses a single type of matches. 
We observed a similar behavior for the Cambridge Landmarks dataset and reported the results in Sec.~\ref{sec:supp_mat}. 

\begin{table*}[t!]
\setlength{\tabcolsep}{3pt}
\begin{center}
\small{
\begin{tabular}{l l l l l l}
 & 1/1 & 1/5 & 1/10 & 1/20 & 1/50 \\
 \cline{2-6}\noalign{\smallskip}
\texttt{P3P}
 & 73.8 / 89.0 / \first{99.0} & 65.4 / 84.3 / 95.3 & 58.6 / \first{79.6} / 89.5 & 34.6 / 57.6 / 69.1 & 9.4 / 20.9 / 38.2 \\
\texttt{E5+1}
& 38.7 / 65.4 / 95.8 & 40.3 / 61.8 / 88.5 & 37.7 / 64.4 / 82.2 & 28.3 / 52.9 / 79.1 & 23.0 / 42.4 / 64.4 \\
\texttt{Adaptive} - mult. msac - hybrid & 70.2 / 88.0 / 98.4 & 63.9 / \first{86.9} / \first{95.8} & 55.0 / 78.0 / \first{90.1} & \first{42.4} / 66.0 / \first{81.7} & \first{23.6} / \first{47.1} / \first{67.5} \\
\texttt{Adaptive} - sum. msac - hybrid & 73.3 / 88.0 / 98.4 & 66.5 / \first{86.9} / \first{95.8} & 60.2 / 79.1 / 89.5 & 40.8 / \first{67.0} / 81.2 & 17.8 / 40.8 / 65.4 \\
\texttt{Select} & 65.4 / 84.8 / 97.9 & 57.6 / 82.2 / 91.6 & 47.1 / 72.3 / 86.4 & 34.6 / 59.2 / 79.1 & 20.4 / 38.7 / 61.3 \\ \hline
\texttt{HLoc}~\cite{sarlin2019coarse,sarlin2020superglue} (no updates) & \first{77.0} / \first{89.5} / \first{99.0} & \first{70.7} / {85.9} / {93.2} & \first{60.7} / 75.4 / 84.3 & 35.1 / 54.5 / 66.5 & 8.4 / 17.3 / 31.9\\
\end{tabular}
}
\end{center}
\caption{Localization with continuous representation updates for the nighttime queries of the Aachen Day-Night v1.1 dataset~\cite{Zhang2020ARXIV, Sattler2018CVPR, Sattler2012BMVC}. The 3D scene structure is represented as an SfM point cloud, and we use every $N$-th database image (1/$N$). 
For reference, we show the results obtained by \texttt{HLoc} without updating the scene representation.}
\label{tab:score_aachen_continuous}
\end{table*}

\begin{table}[t!]
\setlength{\tabcolsep}{3pt}
\begin{center}
\small{
\begin{tabular}{l l l}
 & B1 & B2\\
 \cline{2-3}\noalign{\smallskip}
\texttt{P3P} & 40.4 / 56.6 / 63.9 & 35.7 / 48.3 / 52.1 \\
\texttt{E5+1}\ & 23.9 / 42.4 / 53.1 & 23.9 / 39.8 / 46.4 \\
\texttt{Adaptive} & \first{41.4} / \first{58.1} / \first{65.6} & \first{37.6} / \first{51.0} / \first{54.6} \\
\end{tabular}
}
\caption{Results on the Gangnam Station dataset~\cite{Lee2021LargescaleLD}. We report the percentage of query images within (0.1m, 1°) / (0.25m, 2°) / (1m, 5°) pose errors using an SfM model.
The \texttt{Adaptive} approach (using the multiple of MSAC scores and hybrid local optimization) consistently performs best.}
\label{tab:score_gangnam_static}
\end{center} 
\end{table}

\noindent \textbf{Impact of the scene representation on pose accuracy.} 
The experiments above used an SfM-based scene representation for structure-based pose estimation. 
In this setting, only a subset of all 2D-2D matches has a corresponding 2D-3D match, thus potentially biasing pose estimates to better fit the 2D-2D matches. 
In addition to an SfM-based representation, we also experimented with lifting 2D-3D correspondences from 2D-2D matches using depth maps.
In this case, nearly every 2D-2D match has a corresponding 2D-3D match.
The depth maps were rendered from a MVS (Multi-View Stereo) mesh (obtained from~\cite{Panek2022ECCV}) or NeRF~\cite{Mildenhall2020NeRFRS,nerfstudio} representation or generated using monocular metric depth estimator Metric3D~\cite{yin2023metric, Hu2024Metric3DVA}. 
Results for using a NeRF-based representation are shown in Sec.~\ref{sec:supp_mat}. 
We observe that in some scenes, \texttt{E5+1} outperforms \texttt{P3P} due to inaccuracies in the estimated scene geometry. 
Again, we observe that the \texttt{Adaptive} approach with an MSAC-based scoring function typically offers the best of both pose estimation strategies (structure-based and structure-less).

The depth maps rendered from the mesh result in accurate 2D-3D matches, while the depth maps from the monocular metric depth estimator are noisy and lack consistency between images, resulting in unreliable 2D-3D matches.
The adaptive solver should ideally handle both cases.
This experiment tackles a scenario different from the reducing the set of reference image set (Tab.~\ref{tab:score_aachen}) as the 3D points triangulated from the reduced reference set are sparser, but still accurate. 
Given the amount of noise in the monocular depth maps, we experimented with filtering inaccurate correspondences: 
Every 3D point is projected into all the retrieved reference images containing the corresponding 2D point, and we count the number of such images where the reprojection error is below a threshold (RE thr.).
If the number of such inlier images is smaller than a threshold, the 3D point is filtered out.
In addition to a hard inlier threshold (in. thr.), we also use a second threshold, which is on the fraction of the number of retrieved images observing the corresponding 2D point (rel. thr.).
We compare the scenarios with and without filtering in Tab.~\ref{tab:aachen_depth_exps}.
Filtering can significantly improve the performance of the noisy depth maps from Metric3D~\cite{yin2023metric, Hu2024Metric3DVA}, but can be detrimental if the geometry is accurate.
While the \texttt{Adaptive} method with sum and multiple of MSAC scores performs similarly on SfM representations in Tab.~\ref{tab:score_aachen}, summing the MSAC scores better handles the inaccurate Metric3D depths when combined with filtering in Tab.~\ref{tab:aachen_depth_exps}.
From the experiments, we conclude that \textit{question (C)} can be answered as: \textit{while the choice of structure-based and structure-less approaches can depend on the scene representation, adaptive methods are able to handle different kinds of scene representations out of the box.}
For details, please see Sec.~\ref{sec:supp_mat}.

\PAR{Select strategy.} As can be seen from Tab.~\ref{tab:select}, selecting from two the two pose estimates can outperform \texttt{Adaptive} for the extreme cases where the geometry is either rather unreliable or very reliable.
As can be seen in Tab.~\ref{tab:score_aachen}, in the case where 2D-3D matches remain reliable (but there are fewer of them), there is naturally a benefit when using both matches for pose estimation / refinement.
However, the gap is not drastic for values of $\alpha$ around $0.8$.
Again, this shows that the answer to \textit{question B} is yes, as \textit{the selection strategy is a simple alternative to the adaptive approach}.

\subsection{Practical use in visual localization}
The second set of experiments aims to identify scenarios in which adaptive approaches are especially preferable over structure-based and structure-less methods. 

\noindent \textbf{Sparse scene representations.} 
Previous experiments showed that the main advantage of the \texttt{Adaptive} approach lies in handling sparse scenes where only relatively few database images are used to build the scene representation. 
The same observation can be made for the Extended CMU Seasons dataset in Tab.~\ref{tab:score_cmuse_static}. 
These results are interesting because using a sparser set of database images typically results in more memory-efficient representations. 
In this context, adaptive approaches are an interesting direction to explore. 

\noindent \textbf{Challenging indoor scenes.} Tab.~\ref{tab:score_gangnam_static} shows results on the Gangnam Station dataset~\cite{Lee2021LargescaleLD} capturing a large indoor space with complex appearance (weakly textured surfaces, dynamic video screens, repetitive textures, \etc). 
As can be seen, the adaptive approach consistently improves accuracy.

\noindent \textbf{Continuous scene representation updates.} 
The visual localization literature typically assumes  a static scene representation that is never updated. 
We consider a simple approach to extend the scene representation whenever a query image is localized: 
for a localized image, its pose, its NetVLAD descriptor, its local features, and information about which local feature corresponds to which 3D point in the scene (if any) is stored. 
When localizing a new query image, all original database images and all previously localized queries are considered during retrieval and matching. 
Compared to updating the 3D structure of the scene, \eg, via triangulation and bundle adjustment, this approach is much simpler and much more efficient. 
Note that such an approach can also be used together with a more expensive representation update: 
the latter is run infrequently, \eg, at the end of each day, while the former is run in the meantime.

Tab.~\ref{tab:score_aachen_continuous} shows the results obtained when using this approach to continuously update a scene representation. 
Compared to the results obtained without updates (\cf Tab.~\ref{tab:score_aachen}) and \texttt{HLoc} without updates, the performance drops when using densely sampled database images. 
The drop is particularly noticeable for the \texttt{E5+1} approach, which is very sensitive to the accuracy of the poses of the images in the scene representation. 
The \texttt{P3P} approach is less affected as it does not use the poses of the previously localized queries. 
Still, for sparser representations (using every 20th or 50th image), the \texttt{Adaptive} strategy clearly improves performance, also when compared to the static case. 
This suggests that adaptive approaches can be used to easily densify initially sparse representations.

\section{Conclusion}
\label{sec:conclusion}
In this paper, we have considered the question of whether visual localization approaches that adaptively select between structure-based and structure-less camera pose estimation strategies are useful in practice. 
Our key observation is that the way we select the best camera pose among the poses  estimated by both strategies has a significant impact on pose accuracy and, thus, whether or not an adaptive approach is practically useful. 
Using an appropriate camera pose scoring function and an appropriate local refinement strategy, we have shown that adaptive approaches can offer the best of both worlds: 
high pose accuracy when accurate 3D scene geometry is available (as offered by structure-based methods) and the ability to handle inaccurate scene geometry (as offered by structure-less methods). 
Our experiments show that adaptive approaches are especially useful when only a few database images are available, which is an important scenario when memory consumption is an issue. 
Our approach is simple to implement, and we will make our source code publicly available. 

\noindent\textbf{Acknowledgements. }
This work was supported by
the Czech Science Foundation (GAČR) EXPRO Grant No. 23-07973X and JUNIOR STAR Grant No. 22-23183M,
and the Grant Agency of the Czech Technical University in Prague (No. SGS23/121/OHK3/2T/13).

\begin{appendix}
\section{Supplementary material}
\label{sec:supp_mat}
In this supplementary material, we present visualizations of the NeRF models used in our experiments, the results of the ablation study using the NeRF models as a scene representation, the evaluation on the 7 Scenes dataset~\cite{Glocker2013RealtimeRC, Shotton2013SceneCR}, the ablation on the local optimization approaches, and more details on the setup of the continuous localization experiments.

\subsection*{NeRF scene representation}
We train NeRFs using the Nerfacto~\cite{nerfstudio} model from the Nerfstudio framework~\cite{nerfstudio}.
We selected the model because it offers reasonable training time and the ability to represent even large scenes with a high level of detail.
For outdoor scenes, we use its Huge variant.
Although this model might seem too large for smaller scenes, such as Cambridge Landmarks, we noticed a visible increase in the level of detail.
We use the Base configuration of the Nerfacto model~\cite{nerfstudio} for 7 Scenes, as no significant improvement was observed when using the Huge model configuration as in the case of Cambridge Landmarks.
The model is based on multiresolution hash encoding~\cite{Mller2022InstantNG}, a small MLP, and appearance embedding~\cite{MartinBrualla2020NeRFIT}.
The model also allows for pose refinement~\cite{Wang2021NeRFNR, Lin2020iNeRFIN, Lin2021BARFBN}, which was not used in our experiments, as it resulted in more blurred renderings in our experience.
We use Mask2Former~\cite{cheng2021mask2former, cheng2021maskformer} on outdoor scenes to segment dynamic objects and the sky from the training images.
We train and render the same camera poses, so the NeRF model effectively serves as an implicit representation and depth estimator for the reference image set.

\begin{figure*}[t!]
    \centering
    \includegraphics[width=1.0\textwidth]{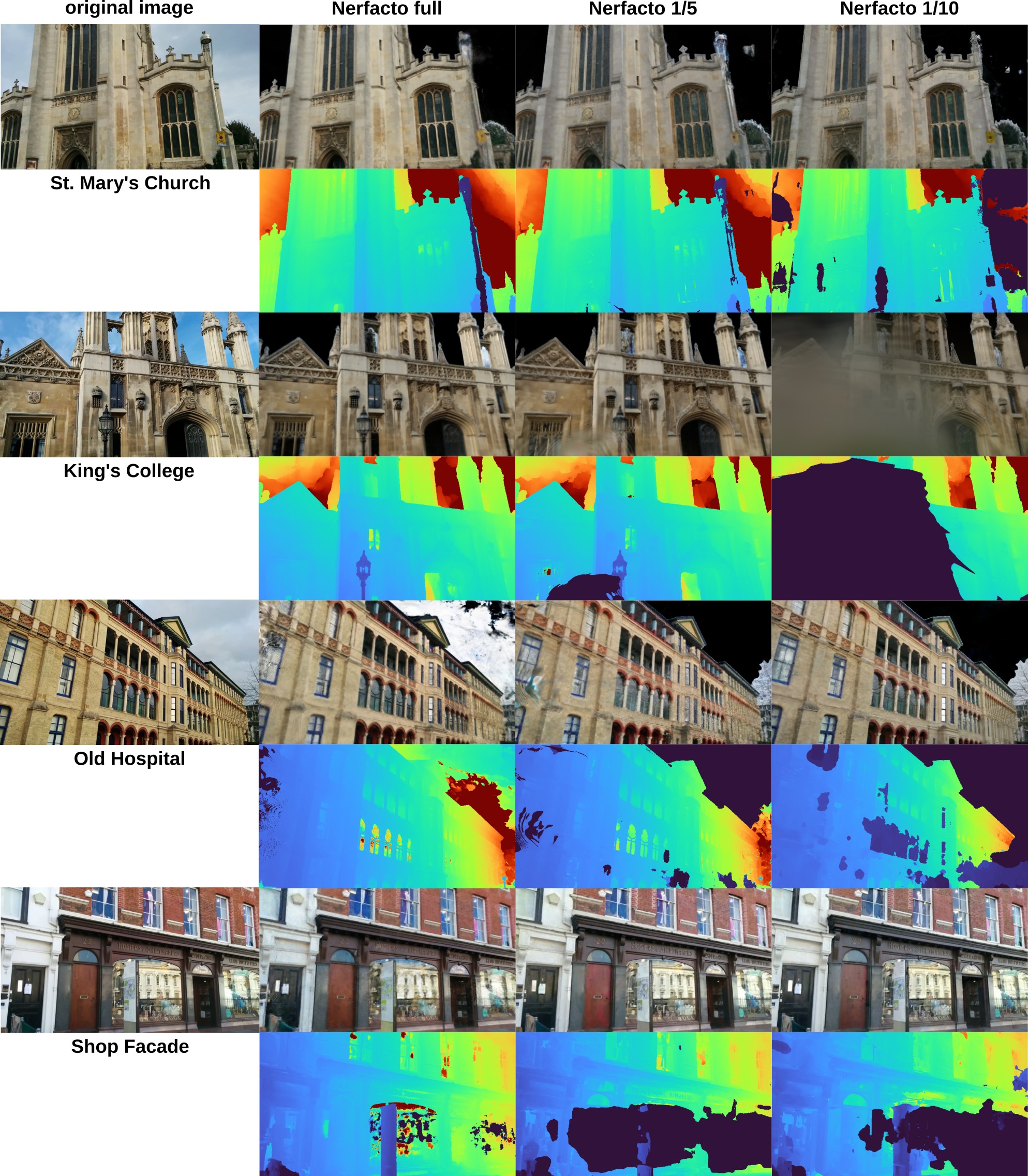}
    \caption{Renderings from Nerfacto Huge model (RGB and depth maps), trained on subsets of the original reference images from the Cambridge Landmarks dataset~\cite{Kendall2015PoseNetAC}. The decreasing number of training images often results in increased floater appearance, which can sometimes completely occlude the view.}
    \label{fig:nerf_comparison_cambridge}
\end{figure*}

When lowering the number of reference images, we, at some point, necessarily encounter degradation of the NeRF model trained on the subset. The lack of training views in combination with purely color-based supervision results in overfitting, creating dense floaters with correct color from the point of view of the training views (\cf the shop windows in Fig.~\ref{fig:nerf_comparison_cambridge}). The presence of the floaters creates patches with wrong values in the rendered depth maps used for the lifting of the 2D-2D matches to 3D (as in MeshLoc~\cite{Panek2022ECCV}).

At localization time, we have RGB renderings and depth maps from the NeRF model.
For every query, we find N most similar rendered RGB images, extract matches between the query and the subset of rendered images, and lift the 2D-2D matches to 3D using the rendered depth maps.
In case the number of reference training, and therefore also the rendered images, is smaller than the selected N (20 in our experiments), the matching is done with all the available reference images.

\begin{figure*}[t!]
    \centering
    \includegraphics[width=1.0\textwidth]{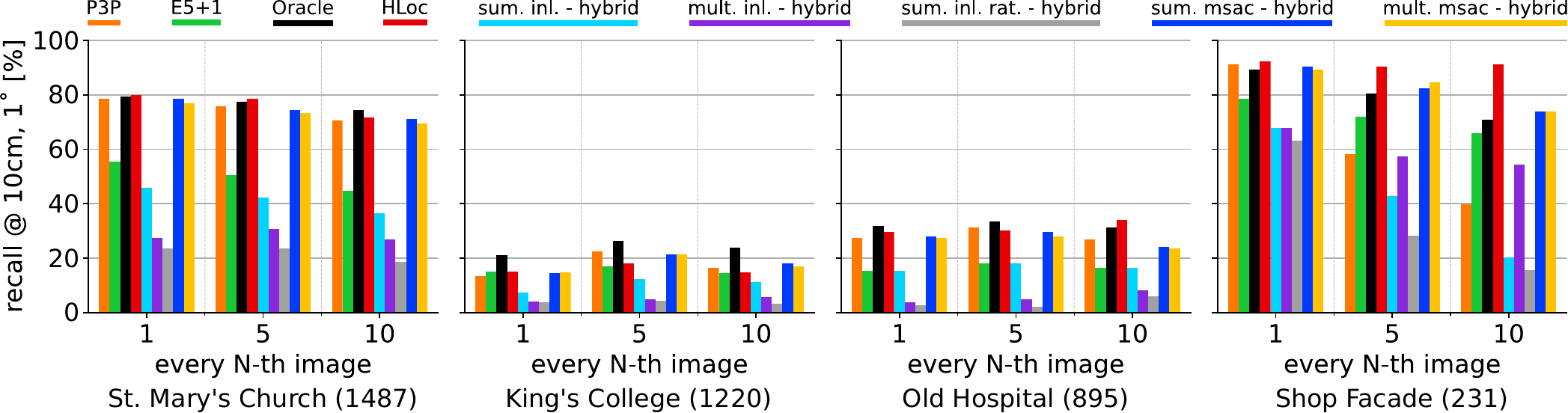}
    \caption{Ablating  scoring functions on the Cambridge Landmarks dataset~\cite{Kendall2015PoseNetAC}. We report the percentage of images localized within 10 cm and 1 degree of the ground truth. The scene is represented using a NeRF trained using every N-th database image.}
    \label{fig:score_cambridge_nerf}
\end{figure*}

\subsection*{Ablation study: scoring functions}
We present the ablation study on the Cambridge Landmarks dataset~\cite{Kendall2015PoseNetAC} using NeRF models, which did not fit into the main paper, in Fig.~\ref{fig:score_cambridge_nerf}.
The results indicate that for some scenes, the \texttt{E5+1} pose solver outperforms \texttt{P3P} because of the inaccuracies in the estimated scene geometry.
The \texttt{Adaptive} approach using the MSAC-based scoring functions achieves the best of structure-based and structure-less pose estimators and is able to handle different kinds of scene representations.

\begin{figure*}[t!]
    \centering
    \includegraphics[width=1.0\textwidth]{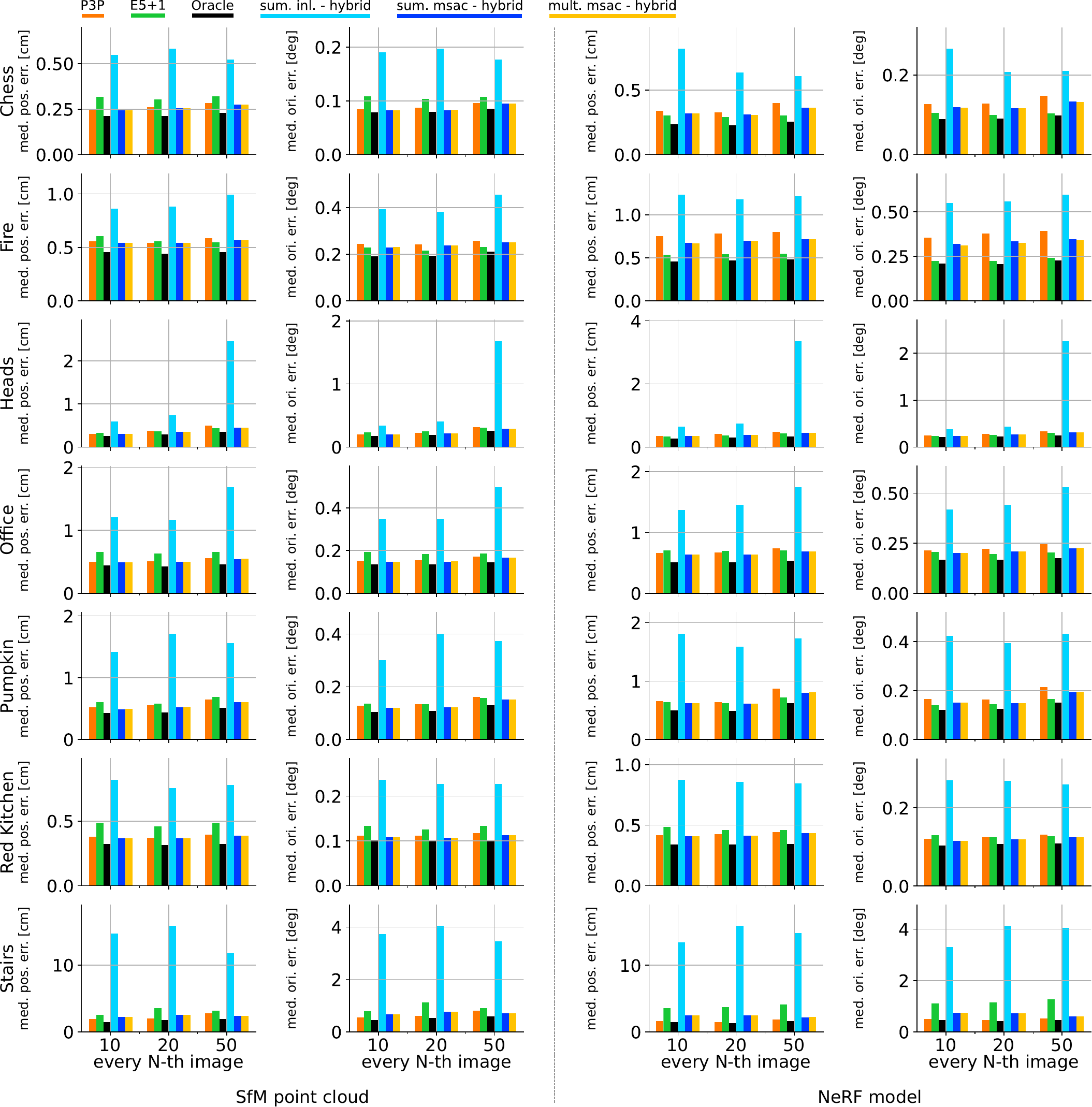}
    \caption{Scoring functions ablation study on the 7 Scenes~\cite{Glocker2013RealtimeRC, Shotton2013SceneCR} dataset. Showing median position and orientation errors (lower is better).}
    \label{fig:seven_median}
\end{figure*}

\begin{figure*}[t!]
    \centering
    \includegraphics[width=1.0\textwidth]{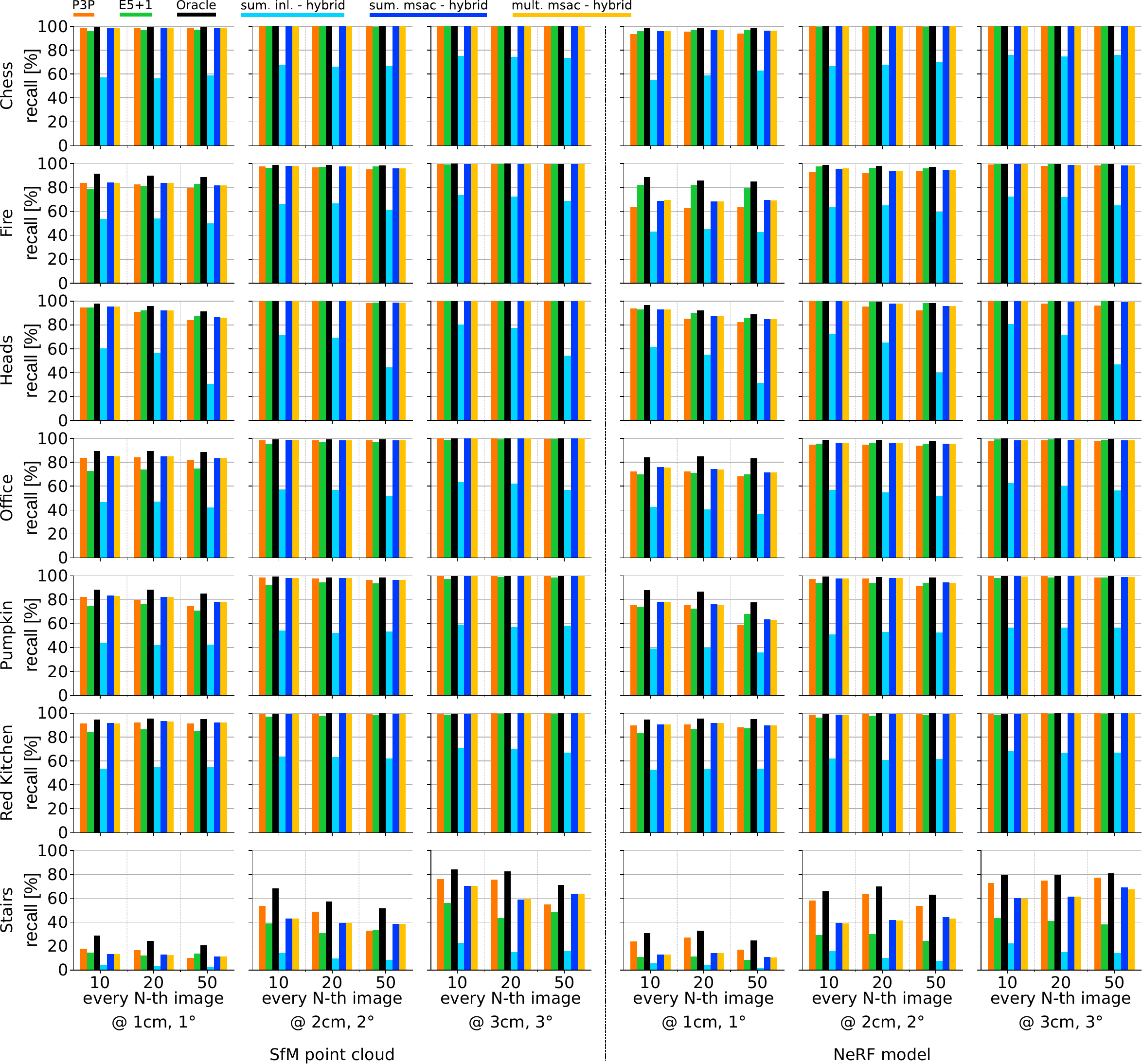}
    \caption{Scoring functions ablation study on the 7 Scenes~\cite{Glocker2013RealtimeRC, Shotton2013SceneCR} dataset. Showing recalls (higher is better) at three different thresholds.}
    \label{fig:seven_recall}
\end{figure*}

The results of the ablation study on the 7 Scenes dataset are presented in Fig.~\ref{fig:seven_median} and Fig.~\ref{fig:seven_recall}.
Tab.~\ref{tab:seven_scenes_img_num} shows the sizes of the references sets of individual scenes from the 7 Scenes dataset.
We follow the same evaluation protocol as used for Cambridge Landmarks~\cite{Kendall2015PoseNetAC} and Aachen Day-Night v1.1~\cite{Zhang2020ARXIV, Sattler2018CVPR, Sattler2012BMVC} in the main paper.
Our initial experiments with NeRFs on the 7 Scenes datasets indicated that the original dataset camera calibration and poses (obtained by KinectFusion~\cite{Izadi2011KinectFusionRD, Izadi2011KinectFusionR3, Newcombe2011KinectFusionRD}) are inaccurate and result in NeRF reconstruction artifacts, such as blurred renderings and view inconsistencies.
Instead we used the SfM-based intrinsics and extrinsics published by Brachmann et al.~\cite{Brachmann2021OnTL} (for both the reference and query images).

We can confirm the major observation from the main paper that the MSAC-based score functions are significantly better than inlier-based scoring functions.
In most of the cases, \texttt{P3P}~\cite{Persson2018ECCV}, \texttt{E5+1}~\cite{Zheng_2015_ICCV}, and the MSAC-based scores have almost identical results.
The first exception is the NeRF-rendered Fire scene at the tightest recall threshold, where \texttt{E5+1} is significantly better, by almost 20 percentage points than \texttt{P3P} and by more than 10 percentage points than the MSAC-based hybrid method.
The most probable reason is the poor geometry learned by the NeRF model, which results in a significant drop in the accuracy of the \texttt{P3P} solver. The low performance of \texttt{P3P} in this particular case also pulls down the performance of the \texttt{Adaptive} approach.

The second exception is the Stairs scene, where \texttt{P3P} is better on the larger subsets of the real images and dominates on the rendered images.
The performance of \texttt{E5+1} is probably skewed by wrong matches on the repetitive grid structure, visible in the scene.
The same issue also influences the results of the hybrid method with MSAC-based scores, which can recover part of the pose precision relative to \texttt{E5+1} but still need to achieve the results of \texttt{P3P}.
Overall, the results on the 7 Scenes are consistent with the results in the main paper.

\begin{table}[t!]
\setlength{\tabcolsep}{3.5pt}
\begin{center}
\small{
\begin{tabular}{l | l l l l l l l}
 & Chess & Fire & Heads & Office & Pumpkin & Red & Stairs \\
 &  &  &  &  &  & Kitchen &  \\ \hline
1/10 & 400 & 200 & 100 & 600 & 400 & 700 & 200 \\
1/20 & 200 & 100 & 50 & 300 & 200 & 350 & 100 \\
1/50 & 80 & 40 & 20 & 120 & 80 & 140 & 40
\end{tabular}
}
\end{center}
\caption{Total numbers of reference images used for our experiments on the 7 Scenes dataset~\cite{Glocker2013RealtimeRC, Shotton2013SceneCR}
when using every $N$-th database image.}
\label{tab:seven_scenes_img_num}
\end{table}

\subsection*{Local optimization}
\begin{figure*}
    \centering
    \includegraphics[width=1.0\textwidth]{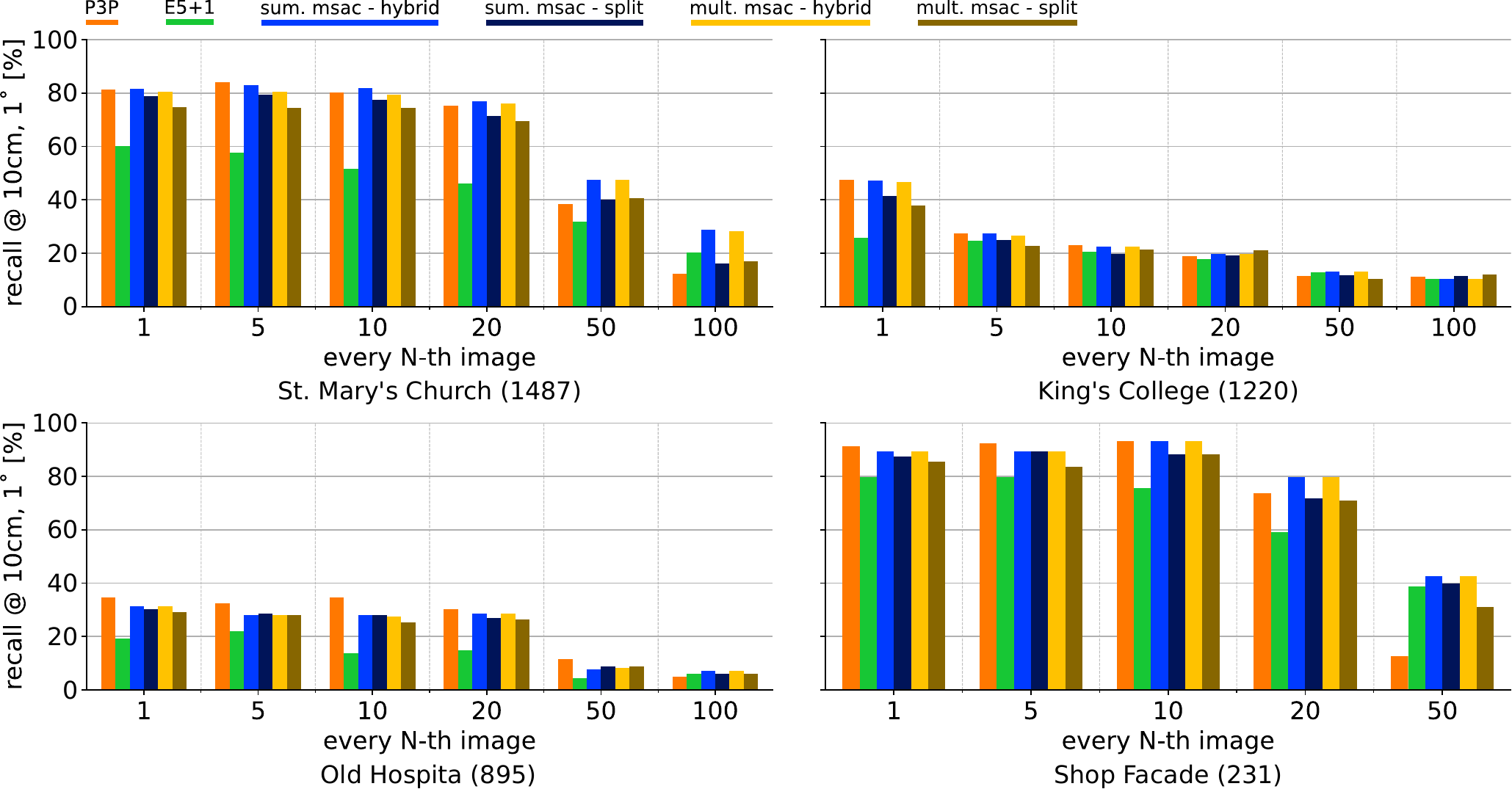}
    \caption{Ablating local optimization approaches on the Cambridge Landmarks dataset~\cite{Kendall2015PoseNetAC}. We report the percentage of images localized within 10 cm and 1 degree of the ground truth. The scene is represented using SfM point clouds computed using every N-th database image.} 
    \label{fig:lo_cambridge_orig}
\end{figure*}

\begin{figure*}
    \centering
    \includegraphics[width=1.0\textwidth]{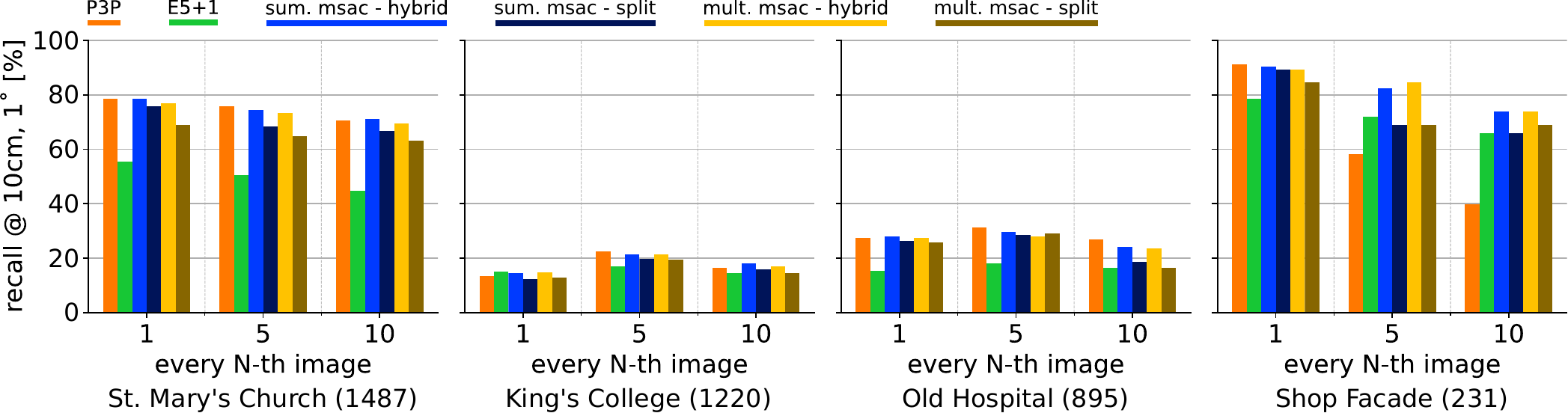}
    \caption{Ablating local optimization approaches on the Cambridge Landmarks dataset~\cite{Kendall2015PoseNetAC}. We report the percentage of images localized within 10 cm and 1 degree of the ground truth. The scene is represented using NeRF trained using every N-th database image.}
    \label{fig:lo_cambridge_nerf}
\end{figure*}

Fig.~\ref{fig:lo_cambridge_orig} and Fig.~\ref{fig:lo_cambridge_nerf} contain the comparison of the two local optimization approaches as described in the Sec.~3 of the main paper. In average the hybrid refinement achieves slightly better results than the split approach.

\subsection*{Continuous localization - experiment setup}
In this section, we add more details to the continuous localization experiments presented in the main paper.
The initial reference SfM model is created using the fixed reference cameras and triangulating SfM point cloud from SuperPoint~\cite{DeTone2017SuperPointSI} matched with LightGlue~\cite{lindenberger2023lightglue}. The queries are ordered by the time of capture, or by their name if the capture time is not available. If the reference set was captured before the queries, the queries are localized from the oldest to the most recent ones. If the reference set was captured in between the query sets (\eg, the case of Extended CMU Seasons), the queries captured before the reference set are processed from the most recent one going back in time, and then the queries captured after the reference set are localized from the oldest going forward in time.
The ordering localizes first the queries most similar to the reference set in case of gradual changes in the scene.

In the case of the Extended CMU Seasons dataset~\cite{Sattler2018CVPR, Badino2011}, we present results on the full reference set and every 11th image subset. The images in the dataset were captured using two cameras. The reason behind selecting every 11th image is that the images are ordered alternately (every odd image belongs to the first camera and every even to the second camera). If every 10th, 20th, or 50th image were selected (as with the other datasets), we would have a subset belonging only to a single camera, which is why we use number 11.

\begin{figure*}[t!]
    \centering
    \includegraphics[width=0.85\textwidth]{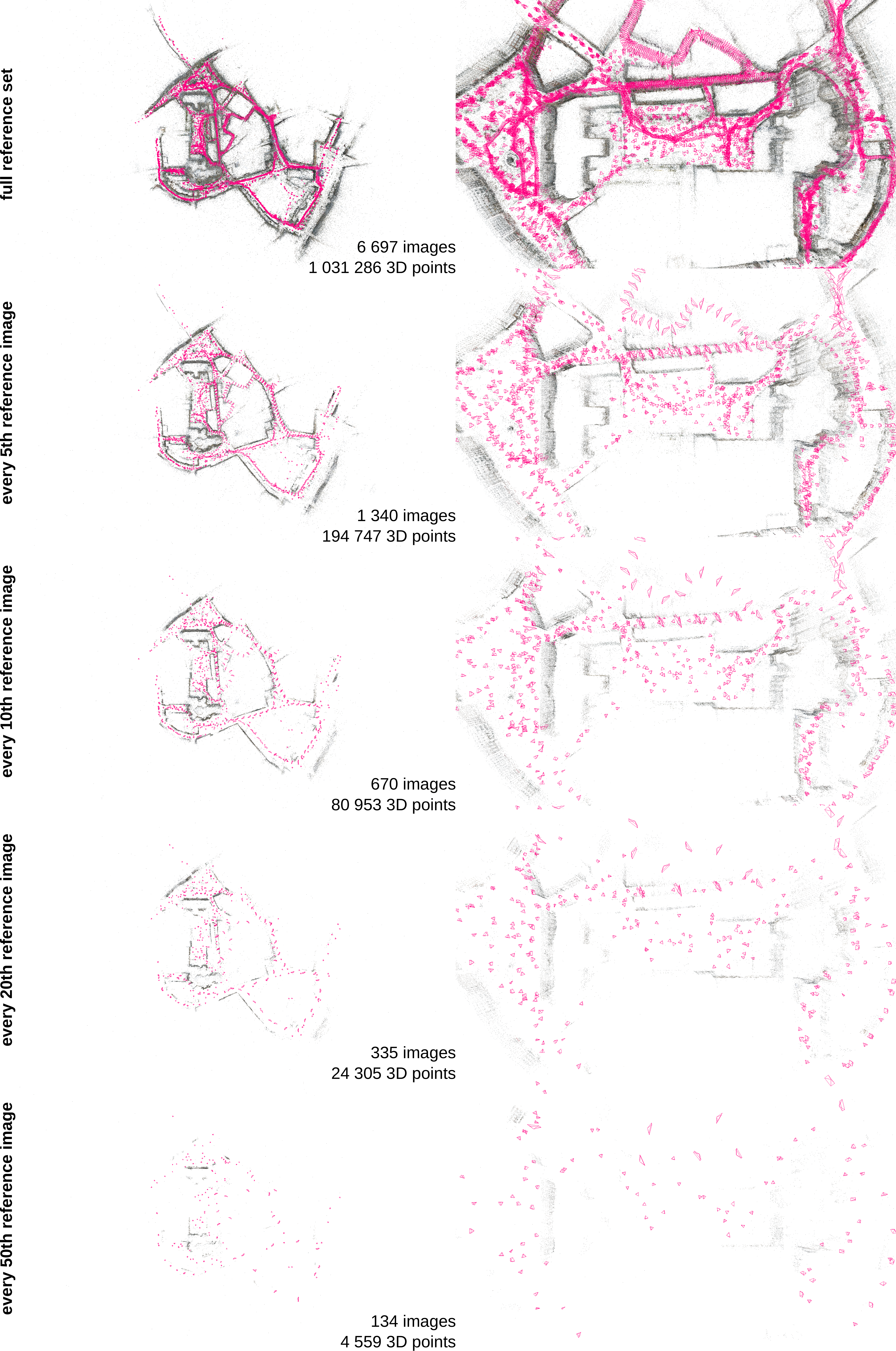}
    \caption{Triangulated models of the Aachen Day-Night v1.1 dataset~\cite{Zhang2020ARXIV, Sattler2018CVPR, Sattler2012BMVC} with different size of reference image set.}
    \label{fig:aachen_ref_models}
\end{figure*}

\begin{figure*}[t!]
    \centering
    \includegraphics[width=1.0\textwidth]{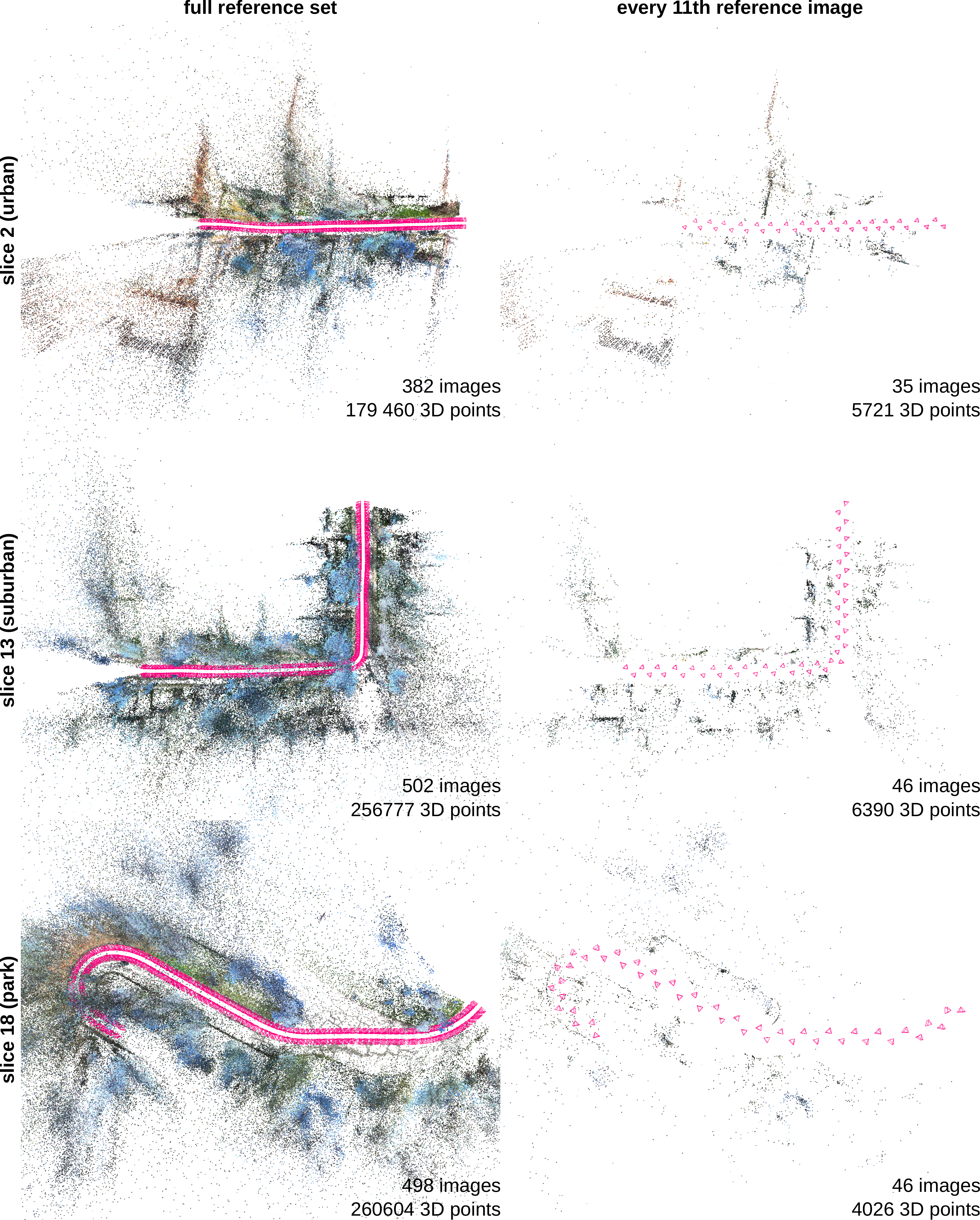}
    \caption{Examples of the triangulated models of the Extended CMU Seasons dataset~\cite{Sattler2018CVPR, Badino2011} used as the initial reference models for the continuous localization experiments.}
    \label{fig:cmuse_ref_models}
\end{figure*}

\begin{figure*}[t!]
    \centering
    \includegraphics[width=1.0\textwidth]{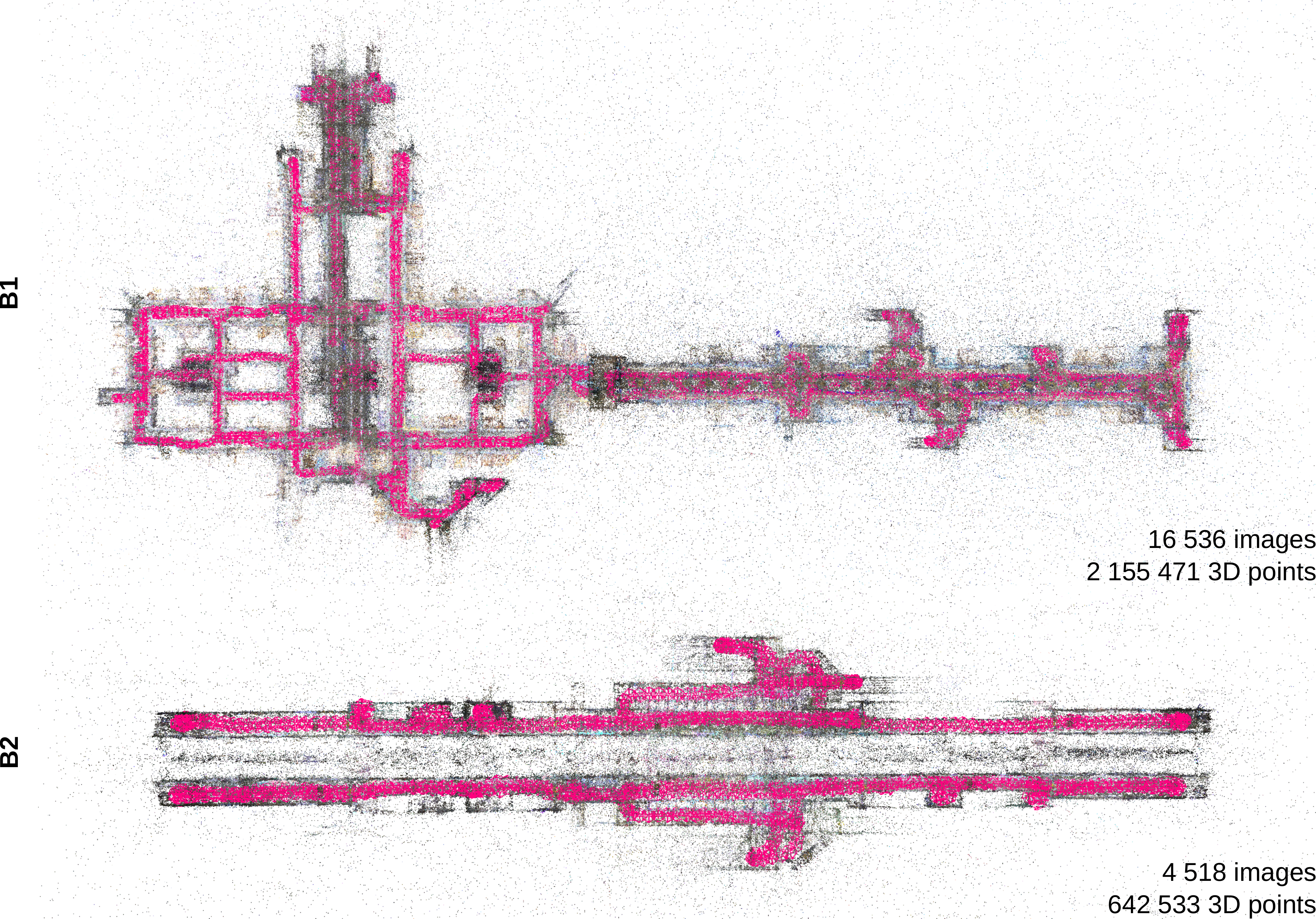}
    \caption{The triangulated models of the NAVER Gangnam Station dataset~\cite{Sattler2018CVPR, Badino2011} used as the initial reference models for the continuous localization experiments.}
    \label{fig:gs_ref_models}
\end{figure*}

We show the examples of initial triangulated reference models and their sizes for the Aachen Day-Night v1.1~\cite{Zhang2020ARXIV, Sattler2018CVPR, Sattler2012BMVC}, Extended CMU Seasons~\cite{Sattler2018CVPR, Badino2011} and Gangnam Station~\cite{Lee2021LargescaleLD} datasets in Fig.~\ref{fig:aachen_ref_models}, Fig.~\ref{fig:cmuse_ref_models} and Fig.~\ref{fig:gs_ref_models}.

\clearpage
\twocolumn

\end{appendix}

{\small
\bibliographystyle{ieee_fullname}
\bibliography{arxiv_bib}
}

\end{document}